\documentclass[10pt,twocolumn,letterpaper]{article}

\usepackage[T1]{fontenc}
\usepackage{cvpr}
\usepackage{times}
\usepackage{epsfig}
\usepackage{graphicx}
\usepackage{amsmath}
\usepackage{amssymb}

\usepackage[pagebackref=true,breaklinks=true,letterpaper=true,colorlinks,bookmarks=false]{hyperref}

\usepackage[square,numbers,sort&compress]{natbib}
\usepackage[inline]{enumitem}
\usepackage{algpseudocode}
\usepackage{array}
\usepackage{multirow}
\usepackage{booktabs}
\usepackage{algorithm}
\usepackage{caption}
\usepackage{subcaption}
\usepackage[normalem]{ulem}
\usepackage{xparse}
\usepackage{pifont}
\usepackage{bm}
\usepackage{listings}
\usepackage{xcolor}

\colorlet{punct}{red!60!black}
\definecolor{background}{HTML}{EEEEEE}
\definecolor{delim}{RGB}{20,105,176}
\colorlet{numb}{magenta!60!black}
\newcommand\BeraMonottfamily{  \def\fvm@Scale{0.85}  \fontfamily{fvm}\selectfont}
\DeclareMathOperator*{\argmax}{argmax}

\newcommand{\redxmark}{{\color{red}\textbf{\ding{55}}}}
\cvprfinalcopy 
\newcommand{\image}{\mathbf{I}}
\newcommand{\imageQpair}{(\mathbf{I}, q)}
\newcommand{\imageQApair}{(\mathbf{I}, q, a)}
\newcommand{\voffline}{v_{\mathrm{offline}}}
\newcommand{\bootstrap}{\mathcal{B}_{\mathrm{init}}}
\newcommand{\trainset}{\mathcal{D}_\mathrm{train}}
\newcommand{\testset}{\mathcal{D}_\mathrm{test}}
\newcommand{\fvalidlabel}{e_\mathrm{valid}}
\newcommand{\fpresentlabel}{e_\mathrm{present}}
\newcommand{\qtype}{\mathrm{q_{type}}}

\newcommand{\clevrtrain}{\texttt{train}\xspace}
\newcommand{\clevrval}{\texttt{val}\xspace}

\newlength\savewidth\newcommand\shline{\noalign{\global\savewidth\arrayrulewidth
  \global\arrayrulewidth 1pt}\hline\noalign{\global\arrayrulewidth\savewidth}}

\newlength\thinwidth\newcommand\thinline{\noalign{\global\savewidth\arrayrulewidth
  \global\arrayrulewidth 0.5pt}\hline\noalign{\global\arrayrulewidth\savewidth}}

\ifcvprfinal\fi
\begin{document}

\title{Learning by Asking Questions}

\author{
    Ishan Misra$^{1}$ \thanks{Work done during internship at Facebook AI Research.}  \quad \quad Ross Girshick$^2$ \quad \quad Rob Fergus$^2$ \\ \quad \quad Martial Hebert$^1$ \quad \quad Abhinav Gupta$^1$ \quad \quad Laurens van der Maaten$^2$ \\
    \vspace{0.1in}
    $^1$Carnegie Mellon University \quad $^2$Facebook AI Research
}

\maketitle

\begin{abstract}

We introduce an interactive learning framework for the development and testing of intelligent visual systems, called learning-by-asking (LBA).  We explore LBA in context of the Visual Question Answering (VQA) task. LBA differs from standard VQA training in that most questions are not observed during training time, and the learner must ask questions it wants answers to. Thus, LBA more closely mimics natural learning and has the potential to be more data-efficient than the traditional VQA setting. We present a model that performs LBA on the CLEVR dataset, and show that it automatically discovers an easy-to-hard curriculum when learning interactively from an oracle. Our LBA generated data consistently matches or outperforms the CLEVR train data and is more sample efficient. We also show that our model asks questions that generalize to state-of-the-art VQA models and to novel test time distributions.
\end{abstract}

\section{Introduction}
Machine learning models have led to remarkable progress in visual recognition. However, while the training data that is fed into these models is crucially important, it is typically treated as predetermined, static information. Our current models are \emph{passive} in nature: they rely on training data curated by humans and have no control over this supervision. This is in stark contrast to the way we humans learn --- by \emph{interacting} with our environment to gain information.
The interactive nature of human learning makes it sample efficient (there is less redundancy during training) and also yields a learning curriculum (we ask for more complex knowledge as we learn).

In this paper, we argue that next-generation recognition systems need to have \emph{agency} --- the ability to decide what information they need and how to get it. We explore this in the context of visual question answering (VQA; \cite{antol15vqa,johnson16clevr,zhu16visual7w}). Instead of training on a fixed, large-scale dataset, we propose an alternative \emph{interactive} VQA setup called \emph{learning-by-asking} (LBA): at training time, the learner receives only images and decides \emph{what questions to ask}. Questions asked by the learner are answered by an oracle (human supervision). At test-time, LBA is evaluated exactly like VQA using well understood metrics.

The interactive nature of LBA requires the learner to construct meta-knowledge about what it knows and to select the supervision it needs. If successful, this facilitates more sample efficient learning than using a fixed dataset, because the learner will not ask redundant questions.

We explore the proposed LBA paradigm in the context of the CLEVR dataset \cite{johnson16clevr}, which is an artificial universe in which the number of unique objects, attributes, and relations are limited. We opt for this synthetic setting because there is little prior work on asking questions about images: CLEVR allows us to perform a controlled study of the algorithms needed for asking questions. We hope to transfer the insights obtained from our study to a real-world setting. 
\begin{figure}[!t]
    \centering
    \includegraphics[width=0.48\textwidth]{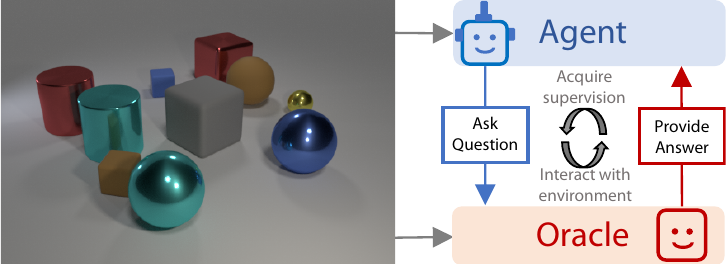}
    \caption{\textbf{The Learning-by-Asking (LBA) paradigm.} We present an open-world Visual Question Answering (VQA) setting in which an agent interactively learns by asking questions to an oracle. Unlike standard VQA training, which assumes a fixed dataset of questions, in LBA the agent has the potential to learn more quickly by asking ``good'' questions, much like a bright student in a class. LBA does not alter the test-time setup of VQA.}
    \label{fig:teaser}
\end{figure}

Building an interactive learner that can ask questions is a challenging task. First, the learner needs to have a ``language'' model to form questions. Second, it needs to understand the input image to ensure the question is relevant and coherent. Finally (and most importantly), in order to be sample efficient, the learner should be able to evaluate its own knowledge (self-evaluate) and ask questions which will help it to learn new information about the world. The only supervision the learner receives from the interaction is the answer to the questions it poses.

We present and study a model for LBA that combines ideas from visually grounded language generation~\cite{mostafazadeh16vqg}, curriculum learning~\cite{bengio2009curriculum}, and VQA. Specifically, we develop an epsilon-greedy~\citep{sutton1998reinforcement} learner that asks questions and uses the corresponding answers to train a standard VQA model. The learner focuses on mastering concepts that it can rapidly improve upon, before moving to new types of questions. We demonstrate that our LBA model not only asks meaningful questions, but also \emph{matches the performance} of human-curated data. Our model is also \emph{sample efficient} and by interactively asking questions it reduces the number of training samples needed to obtain the baseline question-answering accuracy by 40\%.

\vspace{-0.06in}
\section{Related Work}
\vspace{-0.05in}
\textbf{Visual question answering} (VQA) is a surrogate task designed to assess a system's ability to thoroughly understand images. It has gained popularity in recent years due to the release of several benchmark datasets \citep{malinowski14multi,antol15vqa,zhu16visual7w}. Motivated by the well-studied difficulty of analyzing results on real-world VQA datasets \citep{jabri16vqa,zhang16yinyang,ray16question}, Johnson \etal \citep{johnson16clevr} recently proposed a more controlled, synthetic VQA dataset that we adopt in this work.

Current VQA approaches follow a traditional supervised learning paradigm. A large number of image-question-answer triples are collected and a subset of this data is randomly selected for training. Learning-by-asking (LBA) uses an alternative and more challenging setting: training images are drawn from a distribution, but the learner decides what question it needs to ask to learn the most. The learner receives only answer level supervision from these interactions. It must learn to formulate questions as well as model its own knowledge to remove redundancy in question-asking. LBA also has the potential to generalize to open-world scenarios.

There is also significant progress on building models for VQA using LSTMs with convolutional networks \citep{lecun1989backpropagation,hochreiter1997long}, stacked attention networks \citep{yang16stacked}, module networks \citep{andreas16module,johnson17module,hu17learning}, relational networks \citep{santoro17relational}, and others~\citep{perez17film}. LBA is independent of the backbone VQA model and can be used with any existing architecture.

\begin{table}[!t]
\footnotesize{
    \centering
    \begin{tabular}{@{}p{0.23\textwidth}p{0.23\textwidth}@{}}
    \includegraphics[width=0.2 \textwidth]{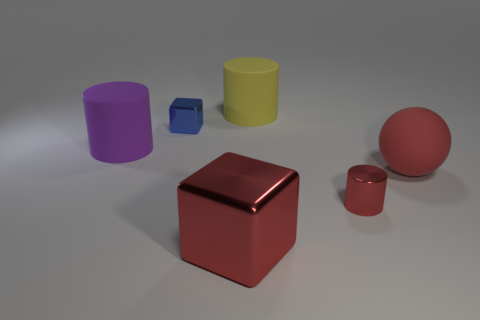}  
         & \includegraphics[width=0.2 \textwidth]{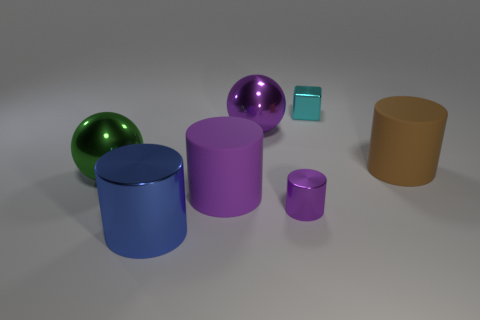} \\
                                \vspace{-0.1in}
        \redxmark\ What size is the purple cube? & \vspace{-0.1in} \redxmark\ What color is the shiny sphere? \\
        \vspace{-0.1in}
        \redxmark\ What size is the red thing in front of the yellow cylinder? & \vspace{-0.1in} \redxmark\ What is the color of the cube to the right of the brown thing? \\
    \end{tabular}
    \vspace{-0.05in}
    \captionof{figure}{Examples of \textbf{invalid} questions for images in the CLEVR universe. Even syntactically correct questions can be invalid for a variety of reasons such as referring to absent objects, incorrect object properties, invalid relationships in the scene or being ambiguous, \etc.}
    \vspace{-0.2in}
    \label{fig:invalid_q_examples}
}
\end{table}

\textbf{Visual question generation} (VQG) was recently proposed as an alternative to image captioning \citep{mostafazadeh16vqg,liu17ivqa,rothe2017question}. Our work is related to VQG in the sense that we require the learner to generate questions about images, however, our objective in doing so is different. Whereas VQG focuses on asking questions that are relevant to the image content, LBA requires the learner to ask questions that are both relevant and informative to the learner when answered. A positive side effect is that LBA circumvents the difficulty of evaluating the quality of generated questions (which also hampers image captioning \citep{anderson16spice}), because the question-answering accuracy of our final model directly correlates with the quality of the questions asked. Such evaluation has also been used in recent works in the language community~\cite{wang2017joint,yang2017semi}.

\textbf{Active learning} (AL) involves a collection of unlabeled examples and a learner that selects which samples will be labeled by an oracle \citep{vijayanarasimhan2014large,kapoor2007active,settles2010active,li2013adaptive}. Common selection criteria include entropy~\cite{joshi2009multi}, boosting the margin for classifiers~\cite{abramson2004active,collins2008towards} and expected informativeness~\cite{houlsby2011bayesian}. Our setting is different from traditional AL settings in multiple ways. First, unlike AL where an agent selects the image to be labeled, in LBA the agent selects an image and \emph{generates a question}. Second, instead of asking for a single image level label, our setting allows for richer questions about objects, relationships \etc for a single image. While \cite{siddiquie2010beyond} did use simple predefined template questions for AL, templates offer limited expressiveness and a rigid query structure. In our approach, questions are generated by a learned language model. Expressive language models, like those used in our work, are likely necessary for generalizing to real-world settings. However, they also introduce a new challenge: there are many ways to generate invalid questions, which the learner must learn to discard (see Figure~\ref{fig:invalid_q_examples}).

\textbf{Exploratory learning} centers on settings in which an agent explores the environment to acquire supervision \citep{storck1995reinforcement}; it has been studied in the context of, among others, computer games and navigation \citep{kulkarni2016hierarchical,pathak2017curiosity}, multi-user games~\citep{merrick2009motivated}, inverse kinematics~\cite{baranes2013active}, and motion planning for humanoids~\citep{frank2014curiosity}. Exploratory learning problems are generally framed with reinforcement learning in which the agent receives (delayed) rewards, which are used to learn a policy that maximizes the expected rewards. A key difference in the LBA setting is that it does \emph{not} have sparse delayed rewards. 
Contextual multi-armed bandits~\citep{bubeck2012regret,langford2008epoch,li2010contextual} are another class of reinforcement learning algorithms that more closely resemble the LBA setting. However, unlike bandits, online performance is irrelevant in our setting: our aim is not to minimize regret, but to minimize the error of the final VQA model produced by the learner.

\begin{figure*}[!t]
    \centering
    \includegraphics[width=\textwidth]{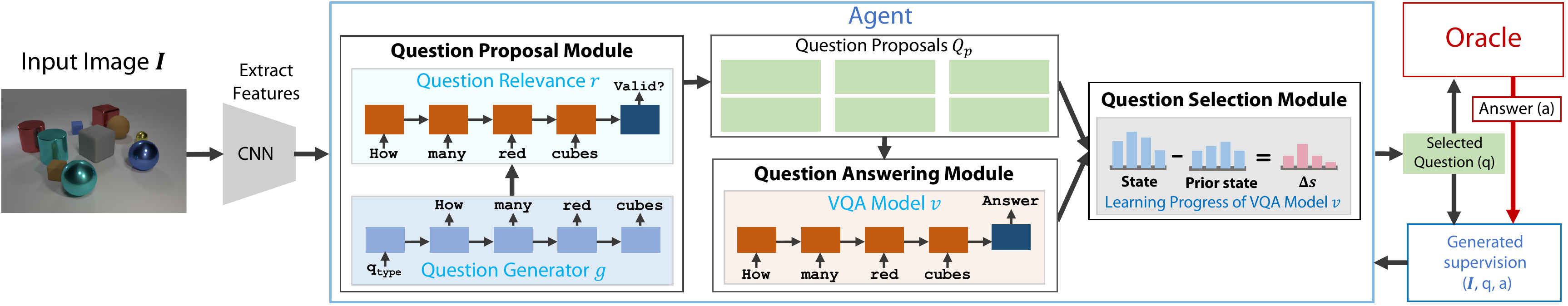}
    \vspace{-0.2in}
    \caption{\textbf{Our approach to the learning-by-asking setting for VQA.} Given an image $\image$, the agent generates a diverse set of questions using a question generator $g$. It then filters out ``irrelevant'' questions using a relevance model $r$ to produce a list of question proposals. The agent then answers its own questions using the VQA model $v$. With these predicted answers and its self-knowledge of past performance, it selects one question from the proposals to be answered by the oracle. The oracle provides answer-level supervision from which the agent learns to ask informative questions in subsequent iterations.}
    \vspace{-0.15in}
    \label{fig:approach}
\end{figure*}

\section{Learning by Asking}

We now formally introduce the learning-by-asking (LBA) setting. We denote an image by $\image$, and assume there exists a set of all possible questions $\mathcal{Q}$ and a set of all possible answers $\mathcal{A}$. At training time, the learner receives as input: (1) a training set of $N$ images, $\trainset \!=\! \{\image_1, \dots, \image_N\}$, sampled from some distribution $p_\mathrm{train}(\image)$; (2) access to an oracle $o(\image, q)$ that outputs an answer $a \in \mathcal{A}$ given a question $q \in \mathcal{Q}$ about image $\image$; and (3) a small bootstrap set of $\imageQApair$ tuples, denoted $\bootstrap$.

The learner receives a budget of $B$ answers that it can request from the oracle. Using these $B$ oracle consultations, the learner aims to construct a function $v(a | \image, q)$ that predicts a score for answer $a$ to question $q$ about image $\image$. The small bootstrap set is provided for the learner to initialize various model components; as we show in our experiments, training on $\bootstrap$ alone yields poor results. 

The challenge of the LBA setting implies that, at training time, \emph{the learner must decide which question to ask about an image} and the only supervision the oracle provides are the answers. As the number of oracle requests is constrained by a budget $B$, the learner must ask questions that maximize (in expectation) the learning signal from each image-question pair sent to the oracle.

At test time, we assume a standard VQA setting and evaluate models by their question-answering accuracy. The agent receives as input $M$ pairs of images and questions, $\testset \!=\! \{(\image_{N+1}, q_{N+1}), \dots, (\image_{N+M}, q_{N+M})\}$, sampled from a distribution $p_\mathrm{test}\imageQpair$. The images in the test set are sampled from the same distribution as those in the training set: $\sum_{q \in \mathcal{Q}} p_\mathrm{test}\imageQpair \!=\! p_\mathrm{train}(\image)$. The agent's goal is to maximize the proportion of test questions that it answers correctly, that is, to maximize:
\begin{equation}
\frac{1}{M} \sum_{m=1}^M \mathbb{I}[\argmax_a v(a | \image_{N+m}, q_{N+m}) = o(\image_{N+m}, q_{N+m})].\nonumber
\end{equation}
We make no assumptions on the marginal distribution over test questions, $p_\mathrm{test}(q)$.

\section{Approach}
\vspace{-0.05in}

We propose a LBA agent built from three modules: (1) a \textbf{question proposal module} that generates a set of question proposals for an input image; (2) a \textbf{question answering module} (or VQA model) that predicts answers from $\imageQpair$ pairs;
and (3) a \textbf{question selection module} that looks at both the answering module's state and the proposal module's questions to pick a single question to ask the oracle.
After receiving the oracle's answer, the agent creates a tuple $\imageQApair$ that is used as the online learning signal for all three modules. Each of the modules is described in a separate subsection below; the interactions between them are illustrated in Figure~\ref{fig:approach}.

For the CLEVR universe, the \textbf{oracle} is a program interpreter that uses the ground-truth scene information to produce answers. As this oracle only understands questions in the form of programs (as opposed to natural language), our question proposal and answering modules both represent questions as programs. However, unlike~\cite{johnson17module,hu17learning}, we do \emph{not} exploit prior knowledge of the CLEVR programming language in any of the modules; instead, it is treated as a simple means that is required to communicate with the oracle. See supplementary material for examples of programs and details on the oracle.

When the LBA model asks an invalid question, the oracle returns a special answer indicating (1) that the question was invalid and (2) whether or not all the objects that appear in the question are present in the image.

\vspace{-0.05in}
\subsection{Question Proposal Module}
\label{sec:question-proposal}
\vspace{-0.05in}

The question proposal module aims to generate a diverse set of questions (programs) that are relevant to a given image. We found that training a single model to meet both these requirements resulted in limited diversity of questions. Thus, we employ two subcomponents: (1) a \textbf{question generation model} $g$ that produces questions $q_{g} \sim g(q | \image)$; and (2) a \textbf{question relevance model} $r(\image, q_{g})$ that predicts whether a generated question $q_{g}$ is \emph{relevant} to an image $\image$. Figure~\ref{fig:invalid_q_examples} shows examples of irrelevant questions that need to be filtered by $r$. The question generation and relevance models are used repeatedly to produce a set of question proposals, $\mathcal{Q}_p \subseteq \mathcal{Q}$.

Our \textbf{question generation model}, $g(q | \image)$, is an image-captioning model that uses a LSTM conditioned on image features (first hidden input) to generate a question. To increase the diversity of generated questions, we also condition the LSTM on the ``question type'' while training~\cite{ferraro16visual} (we use the predefined question types or families from CLEVR). Specifically, we first sample a question type $\qtype$ uniformly at random and then sample a question from the LSTM using a beam size of 1 and a sampling temperature of $1.3$. For each image, we filter out all the questions that have been previously answered by the oracle.

Our \textbf{question relevance model}, $r\imageQpair$, takes the questions from the generator $g$ as input and filters out irrelevant questions to construct a set of question proposals, $\mathcal{Q}_p$. The special answer provided by the oracle whenever an invalid question is asked (as described above) serves as the online learning signal for the relevance model. Specifically, the model is trained to predict (1) whether or not a image-question pair is valid and (2) whether or not all objects that are mentioned in the question are all present in the image. Questions for which both predictions are positive (\emph{i.e.}, that are deemed by the relevance model to be valid and to contain only objects that appear in the image) are put in the question proposal set, $\mathcal{Q}_p$. We sample from the generator until we have 50 question proposals per image that are predicted to be valid by $r\imageQpair$.

\subsection{Question Answering Module (VQA Model)}
\label{sec:question-answering}
Our question answering module is a standard VQA model, $v(a | \image, q)$, that learns to predict the answer $a$ given an image-question pair $\imageQpair$. The answering module is trained online using the supervision signal from the oracle. 

A key requirement for selecting good questions to ask the oracle is the VQA model's capability to self-evaluate its current state. We capture the state of the VQA model at LBA round $t$ by keeping track of the model's question-answering accuracy $s_t(a)$ per answer $a$ on the training data obtained so far. The state captures information on \emph{what the answering module already knows}; it is used by the question selection module.

\vspace{-0.05in}
\subsection{Question Selection Module}
\label{sec:selection-policy}
\vspace{-0.05in}

The question selection module defines a policy, $\pi(\mathcal{Q}_p; \image, s_{1, \dots, t})$, that selects the most informative question to ask the oracle from the set of question proposals $\mathcal{Q}_p$.
To select an informative question, the question selection module uses the current state of the answering module (how well it is learning various concepts) and the difficulty of each of the question proposals. These quantities are obtained from the state $s_t(a)$ and the beliefs of the current VQA model, $v(a | \image, q)$ for an image-question pair, respectively. 

The state $s_t(a)$ contains information about the current knowledge of the answering module. The difference in the state values at the current round, $t$, and a past round, $t-\Delta$, measures how fast the answering module is improving for each answer. Inspired by curriculum learning \citep{bengio2009curriculum,kumar2010self,baranes2013active,sachan2016easy}, we use this difference to select questions on which the answering module can improve the fastest. Specifically, we compute the expected accuracy improvement under the answer distribution for each question $q_p \in \mathcal{Q}_p$:
\begin{equation}
h(q_p; \image, s_{1, \dots, t}) = \sum_{a \in \mathcal{A}} v(a | \image, q_p) \left( \frac{s_{t}(a) - s_{t-\Delta}(a)}{s_{t}(a)} \right). \label{eqn:selection-score}
\end{equation}

We use the expected accuracy improvement as an informativeness value that the learner uses to pick a question that helps it improve rapidly (thereby enforcing a curriculum). In particular, our selection policy, $\pi(\mathcal{Q}_p; \image, s_{1, \dots, t})$, uses the informativeness scores to select the question to ask the oracle using an epsilon-greedy policy~\citep{sutton1998reinforcement}. The greedy part of the selection policy is implemented via $\argmax_{q_p \in \mathcal{Q}_p} h(q_p; \image, s_{1, \dots, t})$, and we set $\epsilon \!=\! 0.1$ to encourage exploration. Empirically, we find that our policy automatically discovers an easy-to-hard curriculum (see Figures \ref{fig:qual_results} and \ref{fig:selection_inform}). In all experiments, we set $\Delta \!=\! 20$; whenever $t \!<\! \Delta$, we set $s_{t - \Delta}(a) \!=\! 0$.

\vspace{-0.05in}
\subsection{Training Phases}
\label{sec:training}
Our model is trained in three phases: (1) an initialization phase in which the generation, relevance, and VQA models ($g$, $r$ and $v$) are pre-trained on a small bootstrap set, $\bootstrap$, of $\imageQApair$ tuples; (2) an online learning-by-asking (LBA) phase in which the model learns by interactively asking questions and updates $r$ and $v$; and (3) an offline phase in which a new VQA model $\voffline$ is trained from scratch on the union of the bootstrap set and all of the $\imageQApair$ tuples obtained by querying the oracle in the online LBA phase.

\par \noindent \textbf{Online LBA training phase.} At each step in the LBA phase (see Figure~\ref{fig:approach}), the proposal module picks an image $\image$ from the training set $\trainset$ uniformly at random.\footnote{A more sophisticated image selection policy may accelerate learning. We did not explore this in our study.}
It then generates a set of relevant question proposals, $\mathcal{Q}_p$ for the image. The answering module tries to answer each question proposal. The selection module uses the state of the answering module along with the answer distributions obtained from evaluating the answering module to pick an informative question, $q$, from the question proposal set. This question is asked to the oracle $o$, which provides just the answer $a = o(\image, q)$ to generate a training example $\imageQApair$. This training example is used to perform a single gradient step on the parameters of the answering module $v$ and the relevance model $r$. The language generation model $g$ remains fixed because the oracle does not provide a direct learning signal for it. This process is repeated until the training budget of $B$ oracle answer requests is exhausted.

\par \noindent \textbf{Offline VQA training phase.} We evaluate the quality of the asked questions by training a VQA model $\voffline$ from scratch on the union of the bootstrap set, $\bootstrap$, and the $\imageQApair$ tuples generated in the LBA phase. We find that offline training of the VQA model leads to slightly improved question-answering accuracy and reduces variance.

\subsection{Implementation Details}
\label{sec:implementation}
The LSTM in $g$ has 512 hidden units. After a linear projection, the image features are fed as its first hidden state. We input a discrete variable representing the question type as the first token into the LSTM before starting generation. Following~\cite{johnson17module}, we use a prefix-tree program representation for the questions.

We implement the relevance model, $r$, and the VQA model, $v$, using the stacked attention network architecture~\citep{yang16stacked} using the implementation of~\citep{johnson17module}. The only modification we make is to concatenate the spatial coordinates to the image features before computing attention as in~\citep{santoro17relational}. We do not share weights between $r$ and $v$.

To generate the invalid pairs $\imageQpair$ for bootstrapping the relevance model, we permute the pairs from the bootstrap set $\bootstrap$ and assume that all such permuted pairs are invalid. Note that the bootstrap set does not have the special answer indicating whether invalid questions ask about objects not present in the image, and these answers are obtained only in the online LBA phase.

Our models use image features from a ResNet-101~\cite{he2016deep} pre-trained on ImageNet~\cite{ILSVRC15}, in particular, from the \texttt{conv4\_23} layer of that network. We use ADAM~\cite{kingma2014adam} with a fixed learning rate of $5e\!-\!4$ to optimize all models.
Additional implementation details are presented in the supplementary material.

\vspace{-0.05in}
\section{Experiments}
\vspace{-0.05in}
\par \noindent \textbf{Datasets.}
We evaluate our LBA approach in the CLEVR universe~\cite{johnson16clevr}, which provides a training set (\clevrtrain) with 70k images and 700k $\imageQApair$ tuples. We use 70k of these tuples as our bootstrap set, $\bootstrap$. We evaluate the quality of the data collected by LBA by measuring the question-answering accuracy of the final VQA model, $\voffline$, on the CLEVR validation (\clevrval)~\cite{johnson16clevr} set. Because CLEVR \clevrtrain and \clevrval have identical answer and question-type distributions, which gives models trained on CLEVR \clevrtrain an inherent advantage. Thus, we also measure question-answering accuracy on the CLEVR-Humans~\cite{johnson17module} dataset, which has a different distribution; see Figure~\ref{fig:clevr_sets_ans_dist}.\footnote{To apply our VQA models to CLEVR-Humans we translate English to CLEVR-programming language using~\cite{johnson17module}; see supplementary material for details.}

\par \noindent \textbf{Models.}
Unless stated otherwise, we use the stacked attention model as the answering module $v$ and evaluate three different choices for the final offline VQA model $\voffline$: 
\par \noindent \textbf{CNN+LSTM} encodes the image using a CNN, the question using an LSTM, and predicts answers using an MLP. 
\par \noindent \textbf{CNN+LSTM+SA} extends CNN+LSTM with the stacked attention (SA) model~\cite{yang16stacked} described in Section~\ref{sec:question-answering}. This is the same as our default answering module $v$.
\par \noindent \textbf{FiLM}~\cite{perez17film} uses question features from a GRU~\cite{cho2014properties} to modulate the image features in each CNN layer.

Unless stated otherwise, we use CNN+LSTM+SA models in all ablation analysis experiments, even though it has lower VQA performance than FiLM, because it trains much faster (6 hours \vs 3 days). For all $\voffline$ models, we use the training hyperparameters from their respective papers.

\subsection{Quality of LBA-Generated Questions}
\label{sec:compare_clevr}

\begin{figure}
    \centering
    \includegraphics[width=0.48\textwidth]{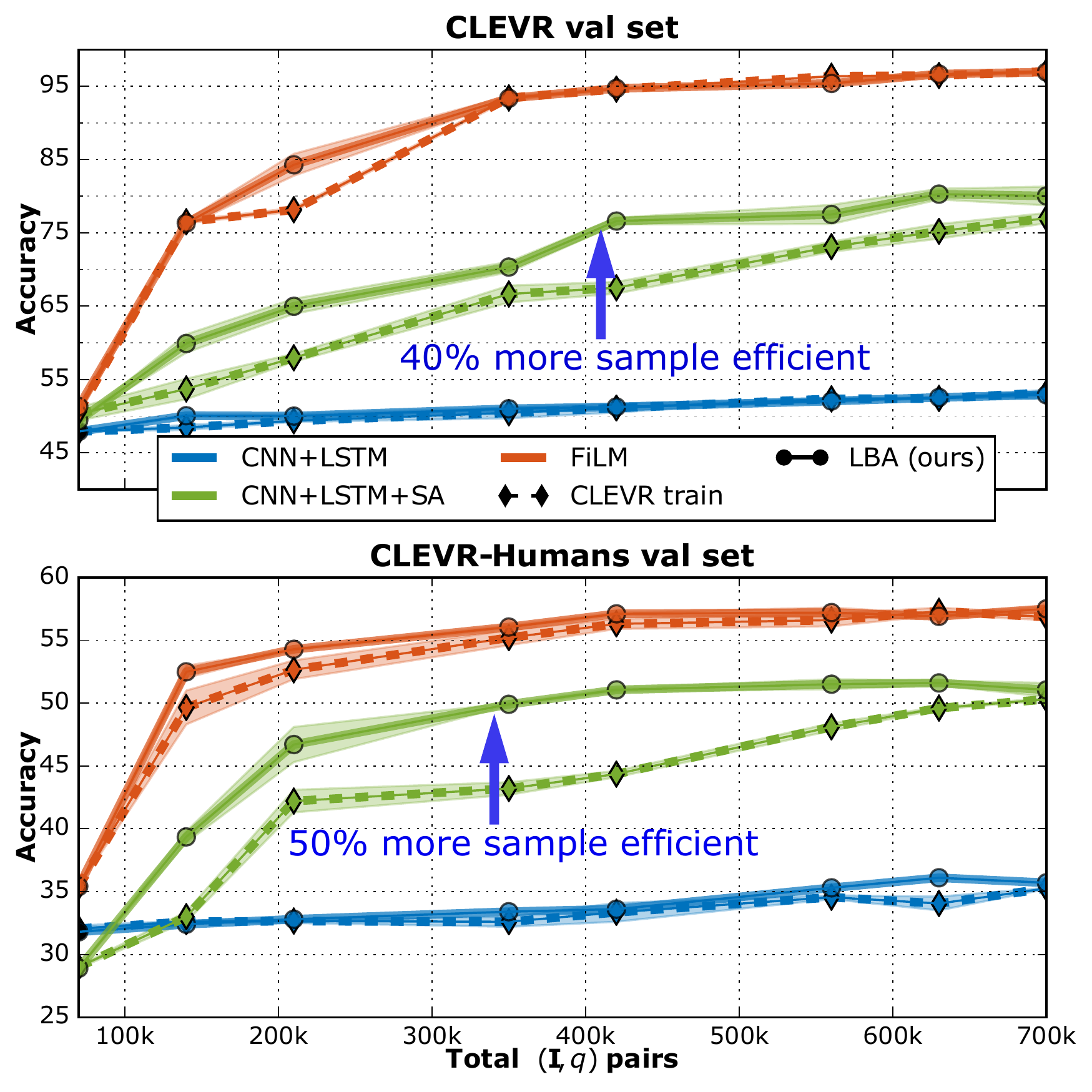}
    \vspace{-0.1in}
    \caption{\textbf{Top:} CLEVR \clevrval accuracy for VQA models trained on CLEVR \clevrtrain (diamonds) \vs LBA-generated data (circles). \textbf{Bottom:} Accuracy on CLEVR-Humans for the same set of models. Shaded regions denote one standard deviation in accuracy. On CLEVR-Humans, LBA is $50\%$ more sample efficient than CLEVR \clevrtrain.}
    \vspace{-0.6em}
    \label{fig:LBA_clevr_cval}
\end{figure}

In Figure~\ref{fig:LBA_clevr_cval}, we compare the quality of the LBA-generated questions to CLEVR \clevrtrain by measuring the question-answering accuracy of VQA models trained on both datasets. The figure shows (top) CLEVR \clevrval accuracy and (bottom) CLEVR-Humans accuracy.
From these plots, we draw four observations.
\par \noindent (1) Using the bootstrap set alone (leftmost point) yields poor accuracy and LBA provides a significant learning signal.
\par \noindent (2) The quality of the LBA-generated training data is at least as good as that of the CLEVR \clevrtrain. This is an impressive result given that CLEVR \clevrtrain has the dual advantage of matching the distribution of CLEVR \clevrval and being human curated for training VQA models. Despite these advantages, LBA matches and sometimes surpasses its performance. More importantly, LBA shows better generalization on CLEVR-Humans which has a different answer distribution (see Figure~\ref{fig:clevr_sets_ans_dist}).
\par \noindent (3) LBA data is sometimes more sample efficient than CLEVR \clevrtrain: for instance, on both CLEVR \clevrval and CLEVR-Humans. The CNN+LSTM+SA model only requires 60\% of $\imageQApair$ LBA tuples to achieve the accuracy of the same model trained on all of CLEVR \clevrtrain.
\par \noindent (4) Finally, we also observe that our LBA agents have low variance at each sampled point during training. The shaded error bars show one standard deviation computed from 5 independent runs using different random seeds. This is an important property for drawing meaningful conclusions from interactive training environments (\cf, \cite{henderson2017deep}).

\par \noindent \textbf{Qualitative results.} Figure~\ref{fig:qual_results} shows five samples from the LBA-generated data at various iterations $t$. They provide insight into the curriculum discovered by our LBA agent. Initially, the model asks simple questions about colors (row 1) and shapes (row 2). It also makes basic mistakes (rightmost column of rows 1 and 2). As the answering module $v$ improves, the selection policy $\pi$ asks more complex questions about spatial relationships and counts (rows 3 and 4).

\begin{table}[!t]
  \setlength\extrarowheight{-3pt}
\centering\footnotesize{
        \begin{tabular}{@{}c@{}}
    \includegraphics[width=0.43    \textwidth]{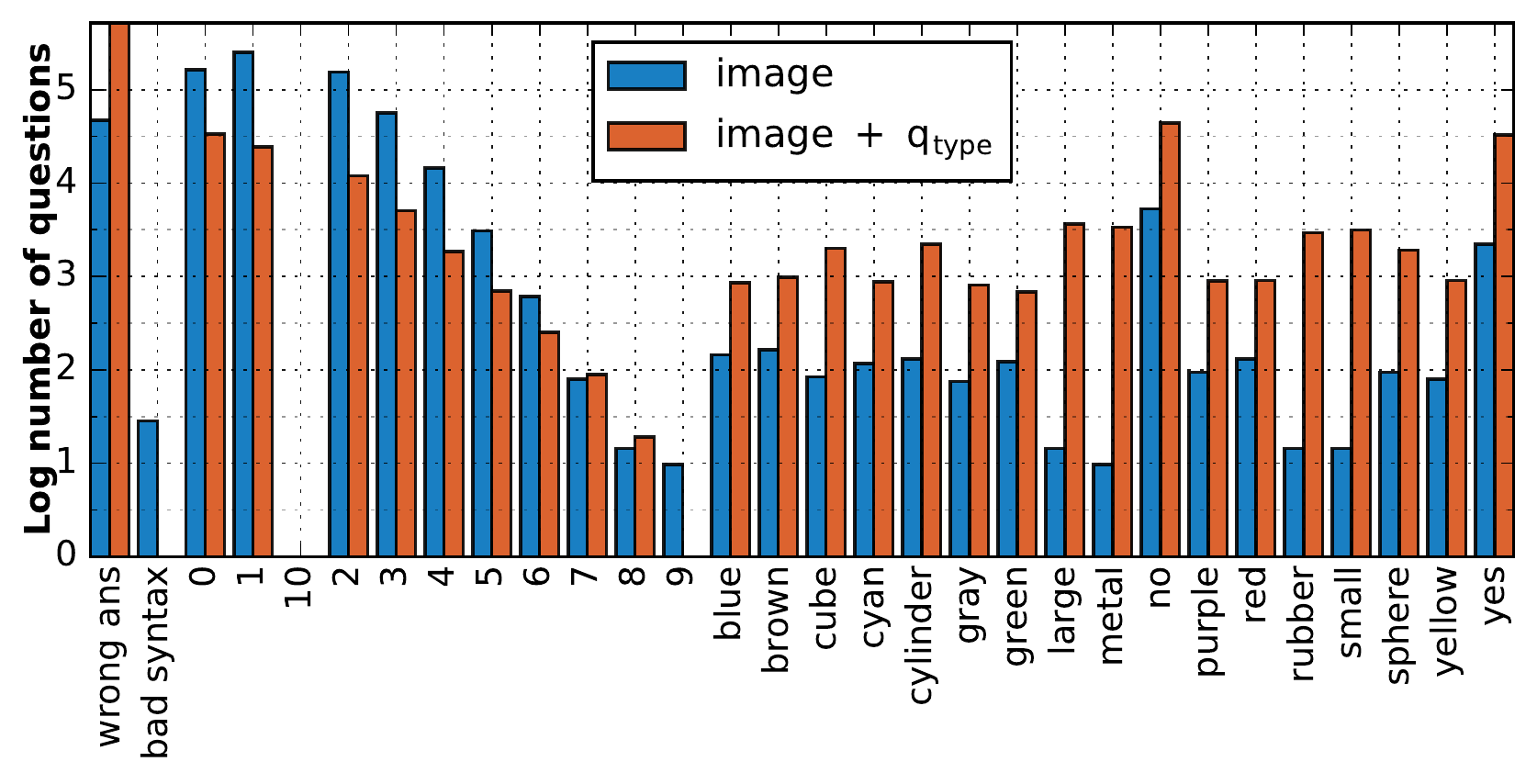}
    \vspace{-0.1in}
    \\
        \includegraphics[width=0.455\textwidth]{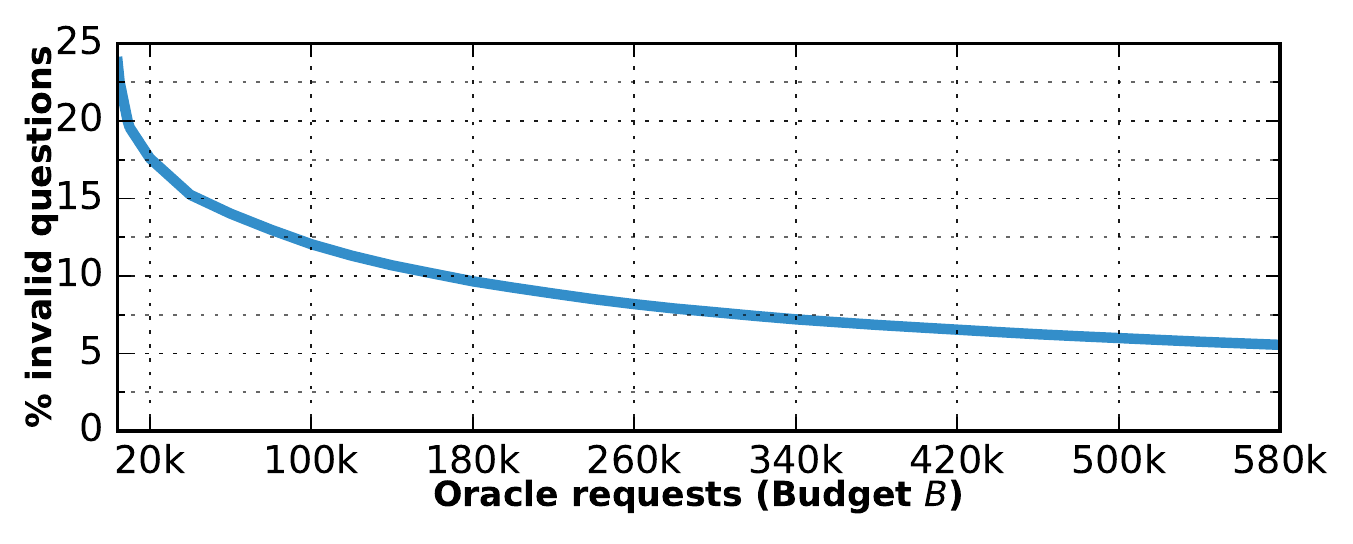}
    \vspace{-0.05in}
            \end{tabular}
    \vspace{-0.15in}
    \captionof{figure}{\textbf{Top:} Histogram of answers to questions generated by $g$ with and without question-type conditioning. \textbf{Bottom:} Percentage of invalid questions sent to the oracle.}
        \vspace{-0.1in}
    \label{fig:q_proposals}
}
\end{table}

\begin{table}[!t]
\setlength{\tabcolsep}{0.25em}
\centering
\footnotesize{
    \begin{tabular}{@{}lc|cccccc@{}}
     & & \multicolumn{6}{c}{\textbf{Budget} $\bm{B}$} \\
    \textbf{Generator} $\bm{g}$ & \textbf{Relevance} $\bm{r}$ & 0k & 70k & 210k & 350k & 560k & 630k \\
    \shline
    $\image$ & None & 49.4 & 43.2 & 45.4 & 49.8 & 52.9 & 54.7 \\
    $\image + \qtype$ & None & 49.4 & 46.3 & 49.5 & 58.7 & 60.5 & 63.4 \\
    \hline
    $\image + \qtype, \tau\!=\!0.3$ & Ours & 49.4 & 60.6 & 67.4 & 70.2 & 70.8 & 70.1 \\
    $\image + \qtype, \tau\!=\!0.7$ & Ours & 49.4 & 60.2 & 70.5 & 76.7 & 77.5 & 77.6 \\
    $\image + \qtype, \tau\!=\!1.3$ & Ours & 49.4 & 60.3 & 71.4 & 76.9 & 79.8 & 78.2 \\
    \hline
    $\image + \qtype$ & Perfect & 49.4 & 67.7 & 75.7 & 80.0 & 81.2 & 81.1
    \end{tabular}
\vspace{-0.5em}
\caption{CLEVR \clevrval accuracy for six budgets $B$. We condition the generator on the image ($\image$) or on the image and the question type ($\image$ + $\qtype$), vary the generator sampling temperatures $\tau$, and use three different relevance models. We re-run the LBA pipeline for each of these settings.}
\label{tab:ablation}
\vspace{-0.5em}
}
\end{table}

\begin{table}[!t]
\setlength{\tabcolsep}{0.25em}
\centering
\footnotesize{
    \begin{tabular}{@{}l|cccccc@{}}
     & \multicolumn{6}{c}{\textbf{Budget} $\bm{B}$} \\
    $\bm{\voffline}$ \textbf{Model} & 0k & 70k & 210k & 350k & 560k & 630k \\
    \shline
    CNN+LSTM & 47.1 & 48.0 & 49.2 & 49.1 & 52.3 & 52.7 \\
    CNN+LSTM+SA & 49.4 & 63.9 & 68.1 & 76.1 & 78.4 & 82.3 \\
    FiLM & 51.2 & 76.2 & 92.9 & 94.8 & 95.2 & 97.3
    \end{tabular}
\caption{CLEVR \clevrval accuracy for three $\voffline$ models when FiLM is used as the online answering module $v$.}
\label{tab:film_vonline}
\vspace{-1em}
}
\end{table}

\begin{table*}
\setlength{\tabcolsep}{3pt}
\footnotesize{
    \centering
\begin{tabular}{@{}cp{0.190000\textwidth}p{0.190000\textwidth}p{0.190000\textwidth}p{0.190000\textwidth}p{0.190000\textwidth}@{}}
\rotatebox[origin=c]{90}{\textbf{Iteration 64k}} &  \raisebox{-0.5\height}{\includegraphics[width=0.180000\textwidth]{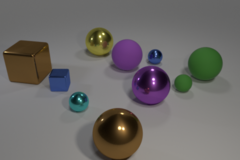}} &  \raisebox{-0.5\height}{\includegraphics[width=0.180000\textwidth]{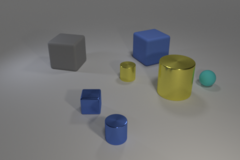}} &  \raisebox{-0.5\height}{\includegraphics[width=0.180000\textwidth]{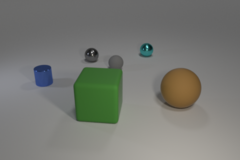}} &  \raisebox{-0.5\height}{\includegraphics[width=0.180000\textwidth]{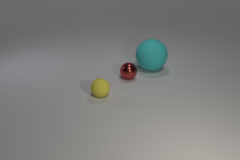}} &  \raisebox{-0.5\height}{\includegraphics[width=0.180000\textwidth]{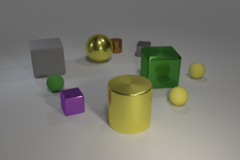}}\\
 &  \textbf{Q:} What is the color of the large metal cube? \textbf{A:} brown &  \textbf{Q:} What is the color of the large cylinder? \textbf{A:} yellow &  \textbf{Q:} What is the color of the small rubber sphere? \textbf{A:} gray &  \textbf{Q:} What is the color of the sphere that is the same size as the red sphere? \textbf{A:} yellow &  \textbf{Q:} What is the color of the object? \textbf{A:} \redxmark \\

 \rotatebox[origin=c]{90}{\textbf{Iteration 256k}} &  \raisebox{-0.5\height}{\includegraphics[width=0.180000\textwidth]{CLEVR_train_001192.png}} &  \raisebox{-0.5\height}{\includegraphics[width=0.180000\textwidth]{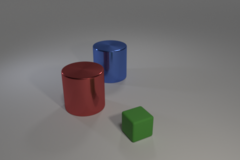}} &  \raisebox{-0.5\height}{\includegraphics[width=0.180000\textwidth]{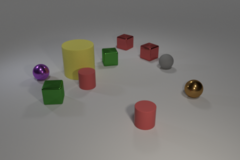}} &  \raisebox{-0.5\height}{\includegraphics[width=0.180000\textwidth]{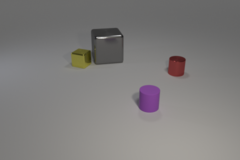}} &  \raisebox{-0.5\height}{\includegraphics[width=0.180000\textwidth]{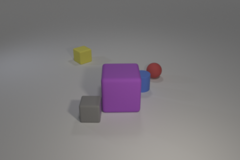}}\\
 &  \textbf{Q:} What is the shape of the yellow object? \textbf{A:} sphere &  \textbf{Q:} What is the shape of the small object? \textbf{A:} cube &  \textbf{Q:} What is the shape of the small brown object? \textbf{A:} sphere &  \textbf{Q:} What is the size of cube to the right of the yellow thing? \textbf{A:} large &  \textbf{Q:} What is the shape of the thing that is the same material as the yellow object? \textbf{A:} \redxmark \\

 \rotatebox[origin=c]{90}{\textbf{Iteration 384k}} &  \raisebox{-0.5\height}{\includegraphics[width=0.180000\textwidth]{CLEVR_train_001192.png}} &  \raisebox{-0.5\height}{\includegraphics[width=0.180000\textwidth]{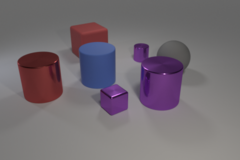}} &  \raisebox{-0.5\height}{\includegraphics[width=0.180000\textwidth]{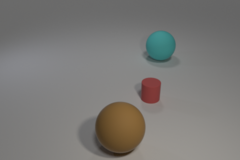}} &  \raisebox{-0.5\height}{\includegraphics[width=0.180000\textwidth]{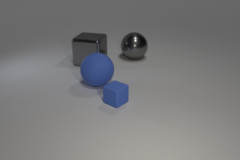}} &  \raisebox{-0.5\height}{\includegraphics[width=0.180000\textwidth]{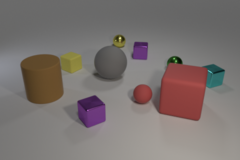}}\\
 &  \textbf{Q:} Is the number of green things greater than the number of brown things? \textbf{A:} no &  \textbf{Q:} Is the number of objects to the left of the small cylinder greater than the number of purple objects? \textbf{A:} yes &  \textbf{Q:} What is the shape of the object to the right of the red thing? \textbf{A:} sphere &  \textbf{Q:} Is the gray sphere of the same material as the large blue object? \textbf{A:} no &  \textbf{Q:} What is the shape of the red rubber object? \textbf{A:} \redxmark \\

 \rotatebox[origin=c]{90}{\textbf{Iteration 576k}} &  \raisebox{-0.5\height}{\includegraphics[width=0.180000\textwidth]{CLEVR_train_001192.png}} &  \raisebox{-0.5\height}{\includegraphics[width=0.180000\textwidth]{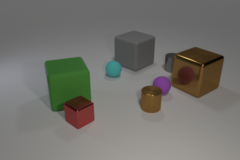}} &  \raisebox{-0.5\height}{\includegraphics[width=0.180000\textwidth]{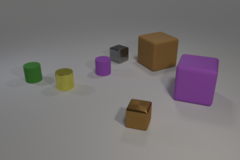}} &  \raisebox{-0.5\height}{\includegraphics[width=0.180000\textwidth]{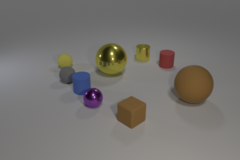}} &  \raisebox{-0.5\height}{\includegraphics[width=0.180000\textwidth]{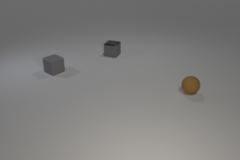}}\\
 &  \textbf{Q:} How many large metal spheres? \textbf{A:} 3 &  \textbf{Q:} Are the number of gray things greater than the number of brown things? \textbf{A:} no &  \textbf{Q:} Is the number of large objects less than the number of cubes? \textbf{A:} yes &  \textbf{Q:} How many objects have the same size as the purple thing? \textbf{A:} 6 &  \textbf{Q:} What is the shape of the brown object to the left of the metal cube? \textbf{A:} \redxmark \\

\end{tabular}
\vspace{-0.1in}
    \captionof{figure}{Example questions asked by our LBA agent at different iterations (manually translated from programs to English). Our agent asks increasingly sophisticated questions as training progresses --- starting with simple color questions and moving on to shape and count questions. We also see that the invalid questions (right column) become increasingly complex.}
\vspace{-0.1in}
    \label{fig:qual_results}
}
\end{table*}

\vspace{-0.05in}
\subsection{Analysis: Question Proposal Module}
\vspace{-0.05in}
\par \noindent \textbf{Analyzing the generator $g$.}
We evaluate the diversity of the generated questions by looking at the distribution of corresponding answers. In Figure~\ref{fig:q_proposals} (top) we use the final LBA model to generate 10 questions for each image in the training set. We plot the histogram of the answers to these questions for generators with and without ``question type'' conditioning. The histogram shows that conditioning the generator $g$ on question type leads to better coverage of the answer space. We also note that about 4\% of the generated questions have invalid programming language syntax.

We observe in the top two rows of Table~\ref{tab:ablation} that the increased question diversity translates into improved question-answering accuracy. Diversity is also controlled by the sampling temperature, $\tau$, used in $g$. Rows 3-5 show that a lower temperature, which gives less diverse question proposals, negatively impacts final accuracy.

\par \noindent \textbf{Analyzing the relevance model $r$.} Figure~\ref{fig:q_proposals} (bottom) displays the percentage of invalid questions sent to the oracle at different time steps during online LBA training. The invalid question rate decreases during training from 25\% to 5\%, even though question complexity appears to be increasing (Figure \ref{fig:qual_results}). This result indicates that the relevance model $r$ improves significantly during training.

We can also decouple the effect of the relevance model $r$ from the rest of our setup by replacing it with a ``perfect'' relevance model (the oracle) that flawlessly filters all invalid questions. Table~\ref{tab:ablation} (row 6) shows that the accuracy and sample efficiency differences between the ``perfect'' relevance model and our relevance model are small, which suggests our model performs well.
\vspace{-0.05in}
\subsection{Analysis: Question Answering Module}
\vspace{-0.05in}
Thus far we have tested our policy $\pi$ with only one type of answering module $v$,  CNN+LSTM+SA. Now, we verify that $\pi$ works with other choices by implementing $v$ as the FiLM model and rerunning LBA. As in Section~\ref{sec:compare_clevr}, we evaluate the LBA-generated questions by training the three $\voffline$ models. The results in Table~\ref{tab:film_vonline} suggest that our selection policy generalizes to a new choice of $v$.

\subsection{Analysis: Question Selection Module}
\label{sec:analyze_selection}
\vspace{-0.05in}
\begin{figure}[!t]
    \centering
    \includegraphics[width=0.48\textwidth]{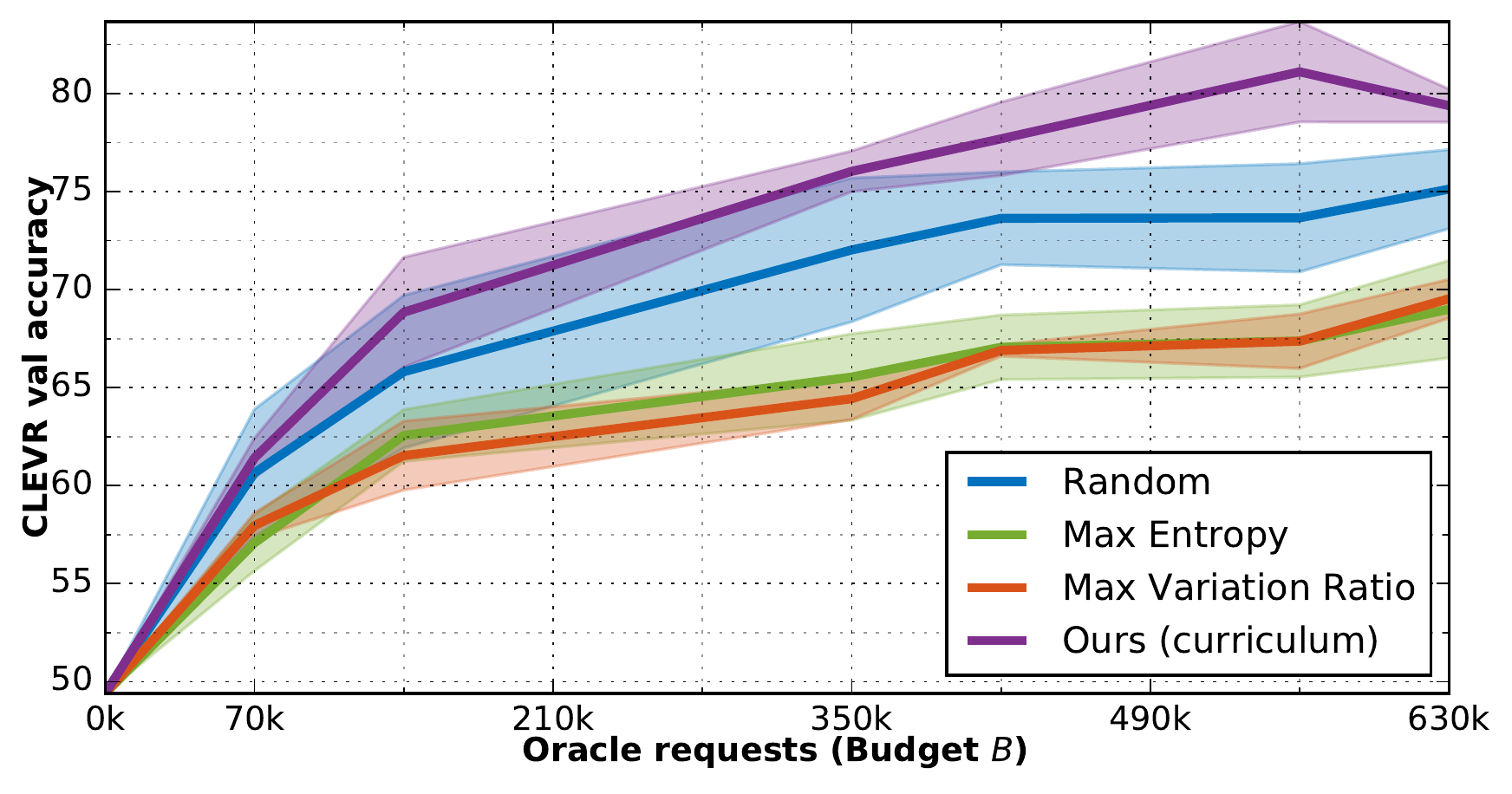}
    \vspace{-0.2in}
    \caption{Accuracy of CNN+LSTM+SA trained using LBA with four different policies for selecting question proposals (Sec~\ref{sec:selection-policy}). Our selection policy is more sample efficient.}
    \vspace{-0.2in}
    \label{fig:ablation_selection}
\end{figure}
\begin{figure}[!h]
    \centering
    \footnotesize{
    \includegraphics[width=0.48\textwidth]{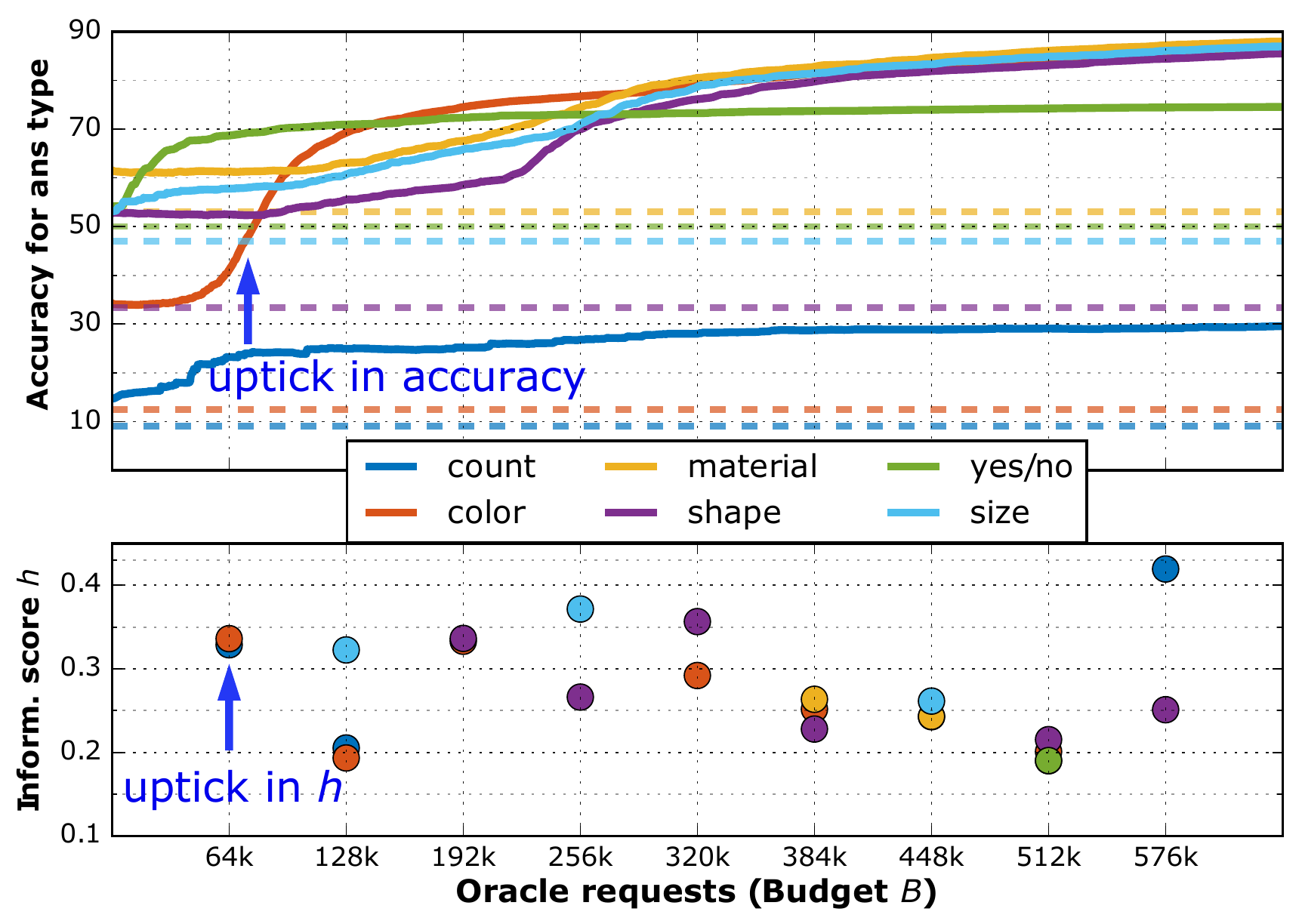}
    \vspace{-0.2in}
    \caption{\textbf{Top:} Accuracy during training (solid lines) and chance level (dashed lines) per answer type. \textbf{Bottom:} Normalized informative scores per answer type, averaged over 10k questions. See Section~\ref{sec:analyze_selection} for details.}
        \label{fig:selection_inform}
    }
\vspace{-1em}
\end{figure}
To investigate the role of the selection policy in LBA, we compare four alternatives: (1) random selection from the question proposals; (2) using the prediction entropy of the answering module $v$ for each proposal after four forward passes with dropout (like in~\cite{sener2017geometric}); (3) using the variation ratio~\cite{freeman1965} of the prediction; and (4) our curriculum policy from Section~\ref{sec:selection-policy}.
We run LBA training with five different random seeds and report the mean accuracy and stdev of a CNN+LSTM+SA model for each selection policy in Figure~\ref{fig:ablation_selection}. In line with results from prior work~\cite{sener2017geometric}, the entropy-based policies perform worse than random selection. By contrast, our curriculum policy substantially outperforms random selection of questions. 
Figure~\ref{fig:selection_inform} plots the normalized informativeness score $h$  (Equation~\ref{eqn:selection-score}) and the training question-answering accuracy ($s(a)$ grouped by per answer type). These plots provide insight into the behavior of the curriculum selection policy, $\pi$. Specifically, we observe a delayed pattern: a peak in the the informativeness score (blue arrow) for an answer type is followed by an uptick in the accuracy (blue arrow) on that answer type. 
We also observe that the policy's informativeness score suggests an easy-to-hard ordering of questions: initially (after 64k requests), the selection policy prefers asking the easier \texttt{color} questions, but it gradually moves on to \texttt{size} and \texttt{shape} questions and, eventually, to the difficult \texttt{count} questions. We emphasize that this easy-to-hard curriculum is learned automatically without any extra supervision.

\subsection{Varying the Size of the Bootstrap Data}
We vary the size of the bootstrap set $\bootstrap$ used for initializing the $g,r,v$ models and analyze its effect on the LBA generated data. In Table~\ref{tab:bootstrap} we show the accuracy of the final $\voffline$ model on CLEVR \clevrval. A smaller bootstrap set results in reduced performance. We also see that with less than $5\%$ (rows 1 and 2) of the CLEVR training dataset as our bootstrap set, LBA asks questions that can match the performance using the entire CLEVR training set. Empirically, we observed that the generator $g$ performs well on smaller bootstrap sets. However, the relevance model $r$ needs enough valid and invalid (permuted) $\imageQApair$ tuples in the bootstrap set to filter irrelevant question proposals. As a result, a smaller bootstrap set affects the sample efficiency of LBA.

\begin{table}[!h]
\setlength{\tabcolsep}{0.4em}
\centering
\footnotesize{
    \begin{tabular}{@{}l|ccccccc@{}}
     & \multicolumn{7}{c}{\textbf{Budget} $\bm{B}$} \\
    $\bm{|\bootstrap|}$ & 0k & 70k & 140k & 210k & 350k & 560k & 630k \\
    \shline
    20k & 48.2 & 56.4 & 63.5 & 66.9 & 72.6 & 75.8 & 76.2 \\
    35k & 48.8 & 58.6 & 64.3 & 68.7 & 74.9 & 76.1 & 76.3 \\
    70k & 49.4 & 61.1 & 67.6 & 72.8 & 78.0 & 78.2 & 79.1 \\
    \end{tabular}
\caption{Accuracy on CLEVR validation data at different budgets $B$ as a function of the bootstrap set size, $|\bootstrap|$.}
\label{tab:bootstrap}
    }
\vspace{-0.5em}
\end{table}

\begin{figure}[!t]
    \centering
    \includegraphics[width=0.45\textwidth]{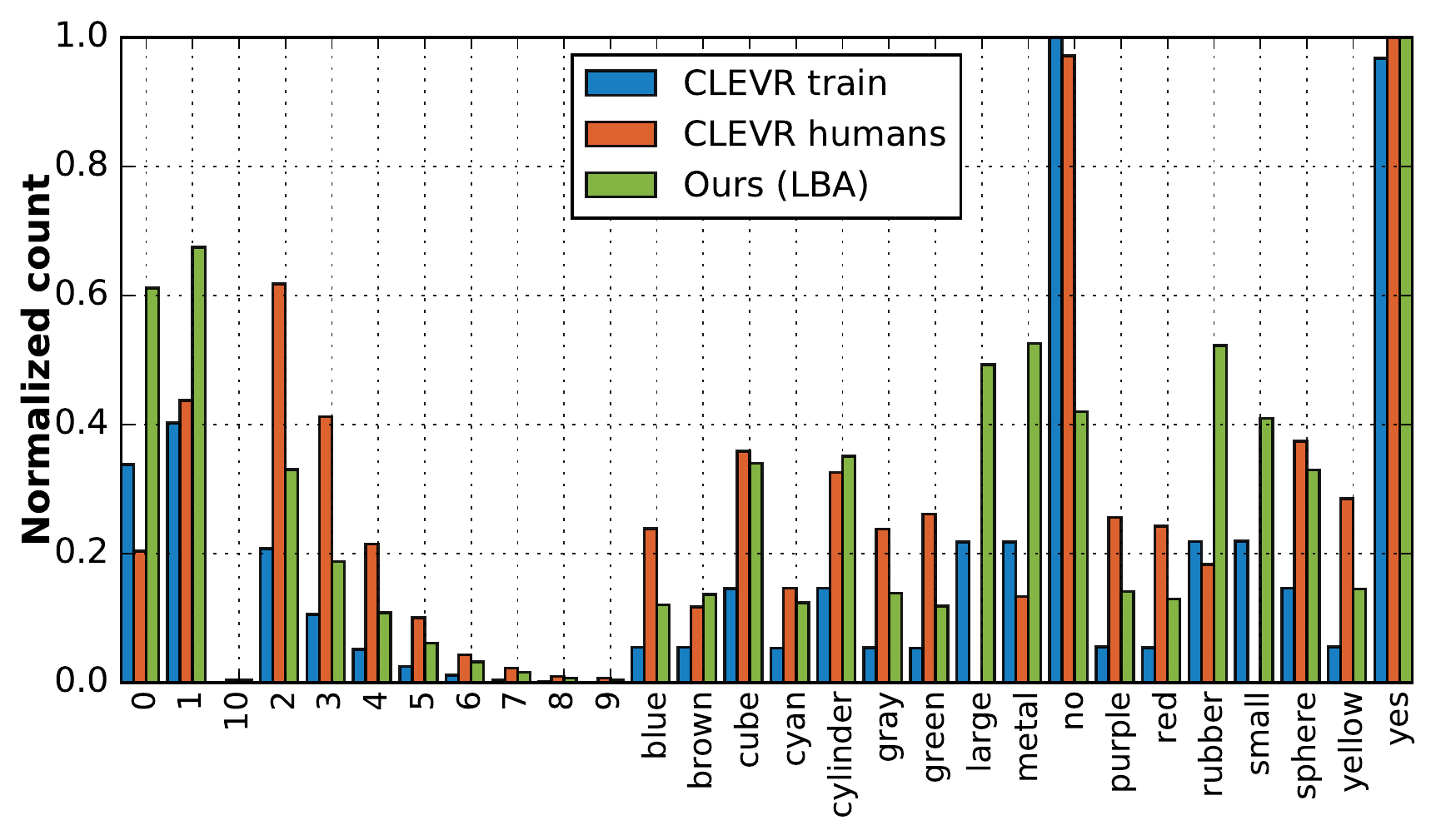}
    \vspace{-0.15in}
    \caption{Answer distribution of CLEVR \clevrtrain, LBA-generated data, and the CLEVR-Humans dataset.}
    \label{fig:clevr_sets_ans_dist}
\vspace{-0.6em}
\end{figure}
\section{Discussion and Future Work}
\vspace{-0.05in}
This paper introduces the learning-by-asking (LBA) paradigm and proposes a model in this setting. LBA moves away from traditional \emph{passively} supervised settings where human annotators provide the training data in an \emph{interactive} setting where the learner seeks out the supervision it needs. While passive supervision has driven progress in visual recognition~\cite{he2016deep,he2017maskrcnn}, it does not appear well suited for general AI tasks such as visual question answering (VQA). Curating large amounts of diverse data which generalizes to a wide variety of questions is a difficult task. Our results suggest that interactive settings such as LBA may facilitate learning with higher sample efficiency. Such high sample efficiency is crucial as we move to increasingly complex visual understanding tasks.

An important property of LBA is that it does not tie the distribution of questions and answers seen at training time to the distribution at test time. This more closely resembles the real-world deployment of VQA systems where the distribution of user-posed questions to the system is unknown and difficult to characterize beforehand~\cite{bottou2013counterfactual}. The CLEVR-Humans distribution in Figure~\ref{fig:clevr_sets_ans_dist} is an example of this. This issue poses clear directions for future work~\cite{bottou2015icml}: we need to develop VQA models that are less sensitive to distributional variations at test time; and not evaluate them under a single test distribution (as in current VQA benchmarks). 

A second major direction for future work is to develop a ``real-world'' version of a LBA system in which (1) CLEVR images are replaced by natural images and (2) the oracle is replaced by a human annotator. Relative to our current approach, several innovations are required to achieve this goal. Most importantly, it requires the design of an effective mode of communication between the learner and the human ``oracle''. In our current approach, the learner uses a simple programming language to query the oracle. A real-world LBA system needs to communicate with humans using diverse natural language. The efficiency of LBA learners may be further improved by letting the oracle return privileged information that does not just answer an image-question pair, but that also explains \emph{why} this is the right or wrong answer~\cite{vapnik2015learning}. We leave the structural design of this privileged information to future work.

\par \noindent \textbf{Acknowledgments:} The authors would like to thank Arthur Szlam, Jason Weston, Saloni Potdar and Abhinav Shrivastava for helpful discussions and feedback on the manuscript; Soumith Chintala and Adam Paszke for their help with PyTorch.

{\small
\bibliographystyle{ieee}
\bibliography{egbib}
}
\newpage
\appendix
\section{Hyperparameters for Models}
\par \noindent \textbf{Generator $g$:} We use 512 hidden units in the LSTM. Each program token is first embedded in a 32-dimensional space before being input to the LSTM. The token embedding is learned jointly with the rest of the model. The generator is trained on the bootstrap set $\bootstrap$ to generate a question $q$ from the image $\image$ by minimizing the cross-entropy loss for generating the question sequence. The question type is a discrete variable that can take one of 90 possible values (the CLEVR datasets defines 90 question families~\cite{johnson16clevr}). The image features are input as the first hidden input. We perform spatial max-pooling on the $1024 \times 14 \times 14$ image features from an ImageNet~\cite{ILSVRC15} pre-trained ResNet-101~\cite{he2016deep} \texttt{conv4\_23} layer to get $1024$ channel features. We then project the features down to 512 dimensions using a linear projection (learned jointly with the model). The model is optimized using SGD with a learning rate of $1.0$ and momentum of $0.9$ with gradients clipped at maximum $L_2$-norm of $5.0$. We use a mini-batch size of 64 and decay the learning rate by 0.25 every $5,000$ iterations. The model is trained for a total of $20,000$ iterations.

\par \noindent \textbf{Relevance model $r$:} We use the stacked-attention network implementation from~\cite{yang16stacked} in our experiments. We use Adam~\cite{kingma2014adam} to minimize the cross-entropy loss of this model, using a fixed learning rate of $5e-4$. We use $L_2$ weight decay of $4e-5$ as a regularizer. The model uses image features from an ImageNet~\cite{ILSVRC15} pre-trained ResNet-101 \texttt{conv4\_23} layer ($1024$ channels). Following~\cite{santoro17relational}, we concatenate spatial coordinates (two channels $x$ and $y$) to these features to finally get $1026$ channel image features of spatial resolution $14 \times 14$.

\par \noindent \textbf{VQA model $v$:} The stacked-attention network serves as the default choice for $v$. We use the same hyperparameters as in the relevance model $r$, but do not share weights between $v$ and $r$.

\section{Details on Oracle}
We use the oracle as implemented in~\cite{johnson16clevr} with the modification that it returns the $\fvalidlabel$ and $\fpresentlabel$ labels in addition to the valid answers. The oracle uses the ground-truth scene graph and relationships provided in CLEVR to answer questions about an image. It takes the program as input and provides the answer. It determines invalid questions as those that do not successfully execute on the scene graph. We show an example of a CLEVR image and its corresponding scene graph in Figure~\ref{fig:scene_graph}.

\section{Extra Experiment: More Questions or More Images?}
Our experiments in Section 5.4 (main paper) show that the LBA setting can be more sample-efficient than a i.i.d. data-collection strategy, \ie, a smaller budget of oracle queries leads to a higher accuracy. We investigate what is more important for learning: the number of images or the number of questions. Specifically, we vary the size of the image set $\trainset$ used for the LBA setting. The results in Table~\ref{tab:vary_images} show that our model can achieve similar results to using the full image set using only 49k of the 70k images. Using 35k images, our model achieves the same accuracy as a model trained on the full CLEVR training set (which has 70k images).

\begin{table}[!t]
\setlength{\tabcolsep}{0.4em}
\centering
\footnotesize{
    \begin{tabular}{@{}l|ccccccc@{}}
     & \multicolumn{7}{c}{\textbf{Budget $B$}} \\
    \textbf{Number of images} & 0k & 70k & 140k & 210k & 350k & 560k & 630k \\
    \shline
    1k & 49.4 & 53.2 & 55.8 & 57.4 & 61.1 & 62.5 & 64.9 \\
    2k & 49.4 & 53.9 & 57.7 & 65.1 & 66.0 & 68.4 & 69.7 \\
    5k & 49.4 & 56.0 & 62.8 & 65.4 & 67.9 & 67.6 & 68.3 \\
    7k & 49.4 & 54.2 & 60.1 & 66.4 & 67.5 & 70.6 & 71.3 \\
    14k & 49.4 & 56.9 & 60.2 & 65.8 & 68.5 & 69.5 &  71.5\\
    21k & 49.4 & 56.8 & 61.5 & 68.1 & 68.7 & 69.8 & 73.4  \\
    35k & 49.4 & 59.7 & 61.6 & 70.6 & 70.8 & 74.3 & 76.6 \\
    49k & 49.4 & 59.5 & 66.9 & 72.7 & 76.7 & 78.1 & 77.2 \\
    70k & 49.4 & 61.1 & 67.6 & 72.8 & 78.0 & 78.2 & 79.1 \\
    \hline
    CLEVR train set (70k) & 49.4 & 55.1 & 57.5 & 65.6 & 72.4 & 74.8 & 77.2  \\
        \end{tabular}
\caption{Accuracy on CLEVR validation data during training as a function of the number of images used.}
\label{tab:vary_images}
}
\end{table}

\section{Translation for CLEVR Humans}
In Section 5.1 of the main paper, we evaluate our models on the CLEVR Humans set. As our models $\voffline$ are trained using programs, we translate the natural language in the CLEVR Humans set using the language-to-program module provided by~\cite{johnson17module}. In Table~\ref{tab:clevr_humans}, we show that such a translation does not affect the accuracy of the models. To do so, we train models on the natural language questions in CLEVR (rather than on the question programs), and evaluate the resulting models on CLEVR-Humans. We compare them with models trained on the CLEVR question programs (like in the main paper), which we evaluate on the translated~\cite{johnson17module}. The results in the table show that both models perform similarly and, therefore, which shows that the language-to-program translation does not impact question-answering accuracy.

\begin{table}[!t]
\setlength{\tabcolsep}{0.4em}
\centering
\footnotesize{
    \begin{tabular}{@{}ll|c@{}}
    \bf Language & \bf Model & \bf Accuracy \\
    \shline
    NLP & CNN+LSTM+SA & 50.1 \\
    Programs & CNN+LSTM+SA & 50.2 \\
    \thinline
    NLP & FiLM & 56.4 \\
    Program & FiLM & 56.3 \\
    \end{tabular}
    \caption{Accuracy of stacked attention networks (CNN+LSTM+SA) and FiLM on the CLEVR Humans validation set. We show that testing the program on either natural language or a program translated version gives similar accuracy.}
\label{tab:clevr_humans}
 }
\end{table}

\begin{table*}
\centering
\begin{tabular}{p{0.5\textwidth}p{0.5\textwidth}}
\captionof{figure}{Example of an image and the associated scene graph.}
\label{fig:scene_graph}
\includegraphics[width=0.45\textwidth]{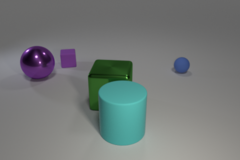} &
\begin{lstlisting}
[
 {
  "color": "green",
  "size": "large",
  "rotation": 194.0002505981195,
  "shape": "cube",
  "3d_coords": [
   1.180546522140503,
   -1.3181802034378052,
   0.699999988079071
  ],
  "material": "metal",
  "pixel_coords": [
   223,
   170,
   9.325896263122559
  ]
 },
 {
  "color": "blue",
  "size": "small",
  "rotation": 155.84760614158418,
  "shape": "sphere",
  "3d_coords": [
   1.1004579067230225,
   2.6546592712402344,
   0.3499999940395355
  ],
  "material": "rubber",
  "pixel_coords": [
   364,
   130,
   11.74298095703125
  ]
 },
 {
  "color": "cyan",
  "size": "large",
  "rotation": 89.26654188937029,
  "shape": "cylinder",
  "3d_coords": [
   2.7460811138153076,
   -2.083714723587036,
   0.699999988079071
  ],
  "material": "rubber",
  "pixel_coords": [
   245,
   223,
   7.810305595397949
  ]
 },
 {
  "color": "purple",
  "size": "large",
  "rotation": 286.7218393410422,
  "shape": "sphere",
  "3d_coords": [
   -2.358253240585327,
   -2.56815242767334,
   0.699999988079071
  ],
  "material": "metal",
  "pixel_coords": [
   76,
   125,
   11.09981632232666
  ]
 },
 {
  "color": "purple",
  "size": "small",
  "rotation": 297.96585798450394,
  "shape": "cube",
  "3d_coords": [
   -2.7632014751434326,
   -0.9295635223388672,
   0.3499999940395355
  ],
  "material": "rubber",
  "pixel_coords": [
   137,
   116,
   12.449520111083984
  ]
 }
]
\end{lstlisting} \\
\end{tabular}
\end{table*}

\newpage

\section{Additional Examples of LBA data}
Following Figure 6 of the main paper, we show a few more \emph{random} examples of the LBA generated data in Figure~\ref{fig:extra_qual}. The programs are translated to natural language manually.

    \begin{table*}
    \setlength{\tabcolsep}{3pt}
    \footnotesize{
        \centering
    \begin{tabular}{@{}cp{0.190000\textwidth}p{0.190000\textwidth}p{0.190000\textwidth}p{0.190000\textwidth}p{0.190000\textwidth}@{}}
\rotatebox[origin=c]{90}{\textbf{Iteration 64k}} &  \raisebox{-0.5\height}{\includegraphics[width=0.180000\textwidth]{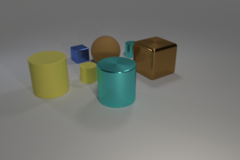}} &  \raisebox{-0.5\height}{\includegraphics[width=0.180000\textwidth]{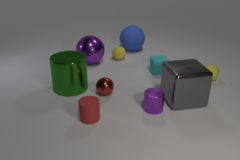}} &  \raisebox{-0.5\height}{\includegraphics[width=0.180000\textwidth]{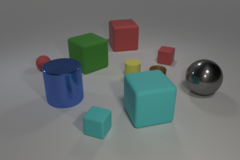}} &  \raisebox{-0.5\height}{\includegraphics[width=0.180000\textwidth]{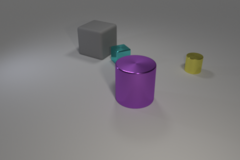}} &  \raisebox{-0.5\height}{\includegraphics[width=0.180000\textwidth]{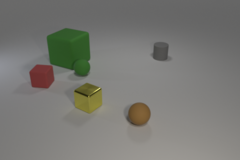}}\\
 &  \textbf{Q:} What is the color of the large metal block? \textbf{A:} brown &  \textbf{Q:} What is the color of the rubber cube? \textbf{A:} cyan &  \textbf{Q:} What is the color of the large metal sphere? \textbf{A:} gray &  \textbf{Q:} What is the color of the large cylinder? \textbf{A:} purple &  \textbf{Q:} What is the color of the small rubber sphere? \textbf{A:} \redxmark \\

 \rotatebox[origin=c]{90}{\textbf{Iteration 128k}} &  \raisebox{-0.5\height}{\includegraphics[width=0.180000\textwidth]{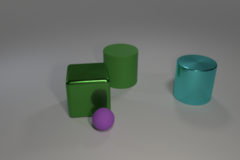}} &  \raisebox{-0.5\height}{\includegraphics[width=0.180000\textwidth]{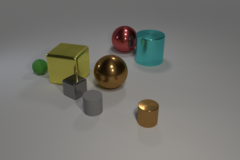}} &  \raisebox{-0.5\height}{\includegraphics[width=0.180000\textwidth]{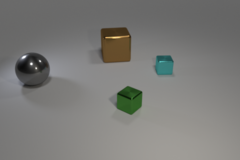}} &  \raisebox{-0.5\height}{\includegraphics[width=0.180000\textwidth]{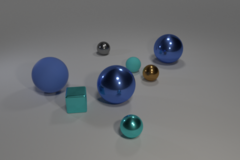}} &  \raisebox{-0.5\height}{\includegraphics[width=0.180000\textwidth]{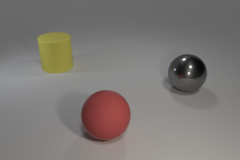}}\\
 &  \textbf{Q:} Is the number of large objects greather than number of purple things? \textbf{A:} yes &  \textbf{Q:} How many large objects? \textbf{A:} 4 &  \textbf{Q:} What is the color of the green cube? \textbf{A:} green &  \textbf{Q:} How many large or large gray spheres? \textbf{A:} 3 &  \textbf{Q:} What is the color of the cube that is the same size as the yellow thing? \textbf{A:} \redxmark \\

 \rotatebox[origin=c]{90}{\textbf{Iteration 256k}} &  \raisebox{-0.5\height}{\includegraphics[width=0.180000\textwidth]{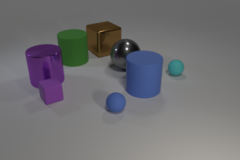}} &  \raisebox{-0.5\height}{\includegraphics[width=0.180000\textwidth]{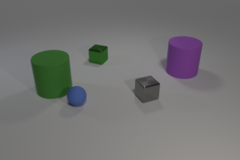}} &  \raisebox{-0.5\height}{\includegraphics[width=0.180000\textwidth]{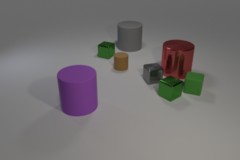}} &  \raisebox{-0.5\height}{\includegraphics[width=0.180000\textwidth]{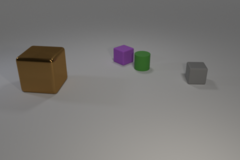}} &  \raisebox{-0.5\height}{\includegraphics[width=0.180000\textwidth]{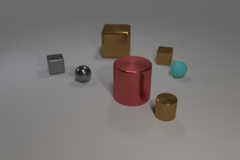}}\\
 &  \textbf{Q:} What is the shape of the large gray object? \textbf{A:} sphere &  \textbf{Q:} What is the size of the green cube? \textbf{A:} small &  \textbf{Q:} What is the size of the purple rubber cylinder? \textbf{A:} large &  \textbf{Q:} Is there an object with the same material as the purple thing? \textbf{A:} yes &  \textbf{Q:} What is the shape of the small gray object? \textbf{A:} \redxmark \\

 \rotatebox[origin=c]{90}{\textbf{Iteration 384k}} &  \raisebox{-0.5\height}{\includegraphics[width=0.180000\textwidth]{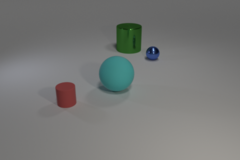}} &  \raisebox{-0.5\height}{\includegraphics[width=0.180000\textwidth]{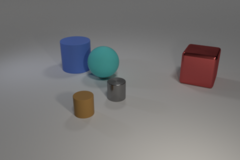}} &  \raisebox{-0.5\height}{\includegraphics[width=0.180000\textwidth]{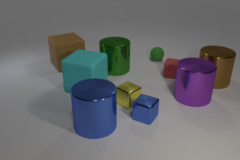}} &  \raisebox{-0.5\height}{\includegraphics[width=0.180000\textwidth]{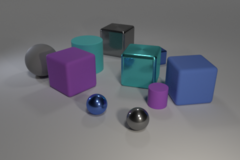}} &  \raisebox{-0.5\height}{\includegraphics[width=0.180000\textwidth]{CLEVR_train_033163.png}}\\
 &  \textbf{Q:} How many objects are to the left or to the right of the small sphere? \textbf{A:} 3 &  \textbf{Q:} What is the shape of the object to the left of the cyan thing? \textbf{A:} cylinder &  \textbf{Q:} What is the shape of the blue cylinder? \textbf{A:} cylinder &  \textbf{Q:} What is the shape of the large gray object? \textbf{A:} sphere &  \textbf{Q:} What is the shape of the small and large purple object? \textbf{A:} \redxmark \\

 \rotatebox[origin=c]{90}{\textbf{Iteration 448k}} &  \raisebox{-0.5\height}{\includegraphics[width=0.180000\textwidth]{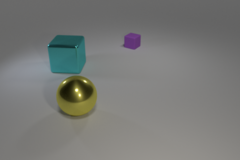}} &  \raisebox{-0.5\height}{\includegraphics[width=0.180000\textwidth]{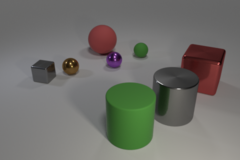}} &  \raisebox{-0.5\height}{\includegraphics[width=0.180000\textwidth]{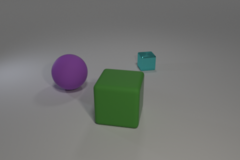}} &  \raisebox{-0.5\height}{\includegraphics[width=0.180000\textwidth]{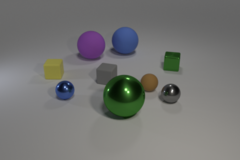}} &  \raisebox{-0.5\height}{\includegraphics[width=0.180000\textwidth]{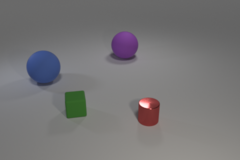}}\\
 &  \textbf{Q:} What is the material of the thing to the right of the cyan cube? \textbf{A:} rubber &  \textbf{Q:} What is the material of the brown sphere? \textbf{A:} metal &  \textbf{Q:} How many objects have the same shape as the green thing? \textbf{A:} 2 &  \textbf{Q:} Is the number of blue things greater than the number of blue things? \textbf{A:} no &  \textbf{Q:} What is the shape of the sphere that is the same material as the green cube? \textbf{A:} \redxmark \\

 \rotatebox[origin=c]{90}{\textbf{Iteration 576k}} &  \raisebox{-0.5\height}{\includegraphics[width=0.180000\textwidth]{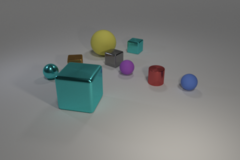}} &  \raisebox{-0.5\height}{\includegraphics[width=0.180000\textwidth]{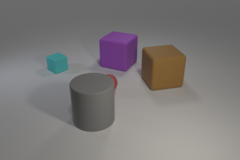}} &  \raisebox{-0.5\height}{\includegraphics[width=0.180000\textwidth]{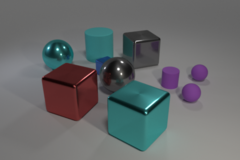}} &  \raisebox{-0.5\height}{\includegraphics[width=0.180000\textwidth]{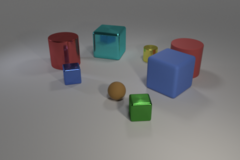}} &  \raisebox{-0.5\height}{\includegraphics[width=0.180000\textwidth]{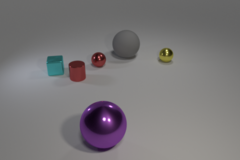}}\\
 &  \textbf{Q:} How many objects are of the same material as the large cube? \textbf{A:} 5 &  \textbf{Q:} How many objects are behind the gray thing and in front of the purple cube? \textbf{A:} 3 &  \textbf{Q:} How many objects have the same material as the large cyan cube? \textbf{A:} 5 &  \textbf{Q:} Is the number of red cylinders less than the blue rubber objects? \textbf{A:} no &  \textbf{Q:} What is the size of the cylinder that is the same size as the small yellow cylinder? \textbf{A:} \redxmark \\

    \end{tabular}
        \captionof{figure}{A few \emph{random} examples of LBA generated data.Our agent asks increasingly sophisticated questions as training progresses -- starting with simple color questions and moving on to shape and count questions. We also see that the invalid questions (right column) become increasingly complex.}
        \label{fig:extra_qual}
    }
    \end{table*}

\section{Examples of CLEVR Programs}
We show a few examples of CLEVR images, questions, programs and answers in Figure~\ref{fig:clevr_ex}. We show examples with short programs for ease of visualization.

        \begin{table*}
        \setlength{\tabcolsep}{3pt}
        \footnotesize{
            \centering
        \begin{tabular}{@{}p{0.190000\textwidth}p{0.190000\textwidth}p{0.190000\textwidth}p{0.190000\textwidth}p{0.190000\textwidth}@{}}
 \raisebox{-0.5\height}{\includegraphics[width=0.180000\textwidth]{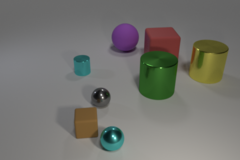}} &  \raisebox{-0.5\height}{\includegraphics[width=0.180000\textwidth]{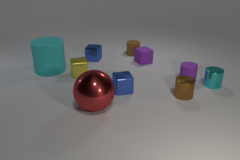}} &  \raisebox{-0.5\height}{\includegraphics[width=0.180000\textwidth]{CLEVR_train_051343.png}} &  \raisebox{-0.5\height}{\includegraphics[width=0.180000\textwidth]{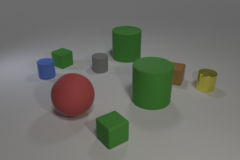}} &  \raisebox{-0.5\height}{\includegraphics[width=0.180000\textwidth]{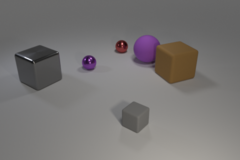}} \\ 
\begin{lstlisting}
[
 {
  "function": "scene", 
  "inputs": [], 
  "value_inputs": []
 }, 
 {
  "function": "filter_size", 
  "inputs": [
   0
  ], 
  "value_inputs": [
   "small"
  ]
 }, 
 {
  "function": "filter_material", 
  "inputs": [
   1
  ], 
  "value_inputs": [
   "rubber"
  ]
 }, 
 {
  "function": "unique", 
  "inputs": [
   2
  ], 
  "value_inputs": []
 }, 
 {
  "function": "query_color", 
  "inputs": [
   3
  ], 
  "value_inputs": []
 }
]\end{lstlisting} & \begin{lstlisting}
[
 {
  "function": "scene", 
  "inputs": [], 
  "value_inputs": []
 }, 
 {
  "function": "filter_shape", 
  "inputs": [
   0
  ], 
  "value_inputs": [
   "sphere"
  ]
 }, 
 {
  "function": "unique", 
  "inputs": [
   1
  ], 
  "value_inputs": []
 }, 
 {
  "function": "query_color", 
  "inputs": [
   2
  ], 
  "value_inputs": []
 }
]\end{lstlisting} & \begin{lstlisting}
[
 {
  "function": "scene", 
  "inputs": [], 
  "value_inputs": []
 }, 
 {
  "function": "filter_size", 
  "inputs": [
   0
  ], 
  "value_inputs": [
   "small"
  ]
 }, 
 {
  "function": "filter_shape", 
  "inputs": [
   1
  ], 
  "value_inputs": [
   "sphere"
  ]
 }, 
 {
  "function": "unique", 
  "inputs": [
   2
  ], 
  "value_inputs": []
 }, 
 {
  "function": "query_color", 
  "inputs": [
   3
  ], 
  "value_inputs": []
 }
]\end{lstlisting} & \begin{lstlisting}
[
 {
  "function": "scene", 
  "inputs": [], 
  "value_inputs": []
 }, 
 {
  "function": "filter_color", 
  "inputs": [
   0
  ], 
  "value_inputs": [
   "yellow"
  ]
 }, 
 {
  "function": "unique", 
  "inputs": [
   1
  ], 
  "value_inputs": []
 }, 
 {
  "function": "query_material", 
  "inputs": [
   2
  ], 
  "value_inputs": []
 }
]\end{lstlisting} & \begin{lstlisting}
[
 {
  "function": "scene", 
  "inputs": [], 
  "value_inputs": []
 }, 
 {
  "function": "filter_size", 
  "inputs": [
   0
  ], 
  "value_inputs": [
   "large"
  ]
 }, 
 {
  "function": "filter_color", 
  "inputs": [
   1
  ], 
  "value_inputs": [
   "green"
  ]
 }, 
 {
  "function": "count", 
  "inputs": [
   2
  ], 
  "value_inputs": []
 }
]\end{lstlisting} \\ 
\textbf{Q:} What color is the tiny rubber thing? & \textbf{Q:} What color is the ball? & \textbf{Q:} The small sphere is what color? & \textbf{Q:} What is the yellow object made of? & \textbf{Q:} How many large green things are there? \\ 
\textbf{A:} brown & \textbf{A:} red & \textbf{A:} blue & \textbf{A:} metal & \textbf{A:} 0 \\

        \end{tabular}
            \captionof{figure}{Examples of programs, questions and answers from the CLEVR dataset.}
            \label{fig:clevr_ex}
        }
        \end{table*}

\end{document}